# Rethinking Generalization in American Sign Language Prediction for Edge Devices with Extremely Low Memory Footprint

Aditya Jyoti Paul [1,2],
1. Department of Computer Science and Engineering,
SRM Institute of Science and Technology,
Kattankulathur, Tamil Nadu – 603203, India.

2. Cognitive Applications Research Lab, India.
aditya_jyoti@srmuniv.edu.in

Puranjay Mohan,
Department of Electronics and Communication Engineering,
SRM Institute of Science and Technology,
Kattankulathur, Tamil Nadu – 603203, India.
puranjaymohan_mu@srmuniv.edu.in

Stuti Sehgal [1,2],
1. Department of Computer Science and Engineering,
SRM Institute of Science and Technology,
Kattankulathur, Tamil Nadu – 603203, India.

2. Cognitive Applications Research Lab, India.
ss3537@srmist.edu.in

*Abstract*— **Due to the boom in technical compute in the last few years, the world has seen massive advances in artificially intelligent systems solving diverse real-world problems. But a major roadblock in the ubiquitous acceptance of these models is their enormous computational complexity and memory footprint. Hence efficient architectures and training techniques are required for deployment on extremely low resource inference endpoints. This paper proposes an architecture for detection of alphabets in American Sign Language on an ARM Cortex-M7 microcontroller having just 496 KB of framebuffer RAM. Leveraging parameter quantization is a common technique that might cause varying drops in test accuracy. This paper proposes using interpolation as augmentation amongst other techniques as an efficient method of reducing this drop, which also helps the model generalize well to previously unseen noisy data. The proposed model is about 185 KB post-quantization and inference speed is 20 frames per second.**

## I. Introduction

American Sign language (ASL) is amongst one of the most widely used sign languages in the world and some estimates indicate it is used by anything between 250 and 500 thousand signers in the USA itself. Hand sign languages vary across the world and the British Sign Language differs almost completely from ASL. ASL by itself, has a lot of variations arising out of not only regional differences, but also by the signers' individual styles. ASL is broadly composed of three major classes: fingerspelling (each alphabet has a symbol), word level vocabulary (each word has a symbol) and non-manual features (the signer uses an amalgamation of facial expressions, mouth, tongue and body poses to communicate).

Thus, it is of pivotal importance to make ASL more accessible, and enable the deaf-mute community to be able to easily communicate with non-signers or signers of other sign languages. This problem has been widely worked upon, and a lot of solutions based on transfer learning and novel but computationally heavy architectures have been proposed. The obvious limitation of many of those approaches would be their impracticality in real-world scenarios.

One use case the authors of this research work envision is a tiny camera in the place where one usually finds the crown or knob of a wristwatch, so that if a person wants to communicate with someone mute, it would be as simple as slightly adjusting one's hand, so the camera points towards the other person, and the user can see their hand signs transcribed to text and/or speech in real-time. Smartwatches might also be able to do this, but many users might not require all the features in those devices and might prefer a cheaper alternative. The model proposed in this paper is tiny enough to run on even the smallest and cheapest of microcontrollers on wearable IoT devices.

TinyML is an emerging field of Machine Learning, encapsulating on-device analytics and inference with very low memory and power usage. Recent advances in this field involve the design of new and innovative architectures and working on unique problems that affect us. This research work sheds some light on the practical implications of training an ASL fingerspelling model on the OpenMV H7 microcontroller board.

Another major challenge working in TinyML with convolutional networks (CNNs) is high generalization error, that is they have high training and test accuracy but do not generalize well to images in new scenarios and noisy backgrounds. However, to make ASL recognition ubiquitous and accessible for everyone, it is of immense importance to perform appropriate generalization testing and this paper reports one of the first research studies taking a step in that direction.

The rest of the paper is organized as follows, section II covers the literature review, section III goes over the technical details and constraints of the problem and approach, section IV covers the chosen experimental methodology, section V discusses the observations, findings and some guiding lights for future researchers and section VI concludes the paper and suggests some possible future avenues of progress in this field.

## II. Literature Review

Sign Language is a non-verbal language used by many members of the deaf-mute community and designing a cheap and efficient ASL detecting pipeline has been an active research topic since the past few years. In [1], Dong et al. cited the complexity of the hand signs, self-occlusion of the hand and limited resolution of the cameras and filters as some of the greatest challenges in ASL recognition, and proposed a novel segmented hand configuration, using a depth contrast feature based per-pixel classification. Various feature extraction methods have been proposed and implemented like Scale-invariant Feature Transform (SIFT) [2][3], Wavelet Moments [4], Histogram of Oriented Gradients (HOG) [5] and Gabor Filters [6]. Classifiers like Artificial Neural Networks (ANNs) [7], Supports Vector Machines (SVMs) [8], and Decision Trees (DTs) [9] have been shown to be robust enough for ASL recognition.

However, most edge devices right now can barely be expected to have Kinect sensors like in [1], or Leap Motion Controllers like in [9][10], while being affordable, and cyber-gloves like in [11] are impractical outside a laboratory setting. Many approaches were proposed using transfer learning like [12] leveraging a GoogLeNet pretrained on ILSVRC2012 dataset, and [13] using MobileNet and Inception_V3. These models won't fit the memory requirements of this study. Bheda et al. [14] used a simple deep convolutional neural network and achieved an accuracy of 82.5% on the pre-existing test set and 67% on their self-created test set, and thus there are some avenues of improvement here.

Recent advances in Neural Network design and performance analysis in ARM Cortex-M CPUs [15], and newer architecture improvements like residual binarized networks [16] and memory-optimal direct convolutions have ushered in newer apostles of efficiency in CNNs with TinyML [17]. Some other interesting problems that must be tackled for efficient ASL recognition are high generalization error in machine learning, where the network is limited to a tiny size and cannot memorize the data but has to learn meaningful representations [18], and effects of color constancy errors like white balance disturbances in the image [19], which inspired the authors to consider the effects of image downsizing and using different interpolations for the same to an advantage in this paper.

## III. Technical Details

This section discusses the details of the experimental setup involving the microcontroller, the software used and the datasets.

### A. Hardware Setup

The experimental setup involves edge deployment on an OpenMV H7 development board which has a 32-bit ARM Cortex-M7 based microcontroller STM32H743VI, with 1MB of SRAM, 2MB of flash memory and clock speed of 400MHz. The frame buffer RAM is however only 496 KB and that's used to load the transient images from the camera, the ML model and associated scratch buffer besides pre-existing system processes.

The documentation suggested keeping the model and scratch buffer under 400 KB, which didn't work during this study. The largest TF Lite model successfully loaded was of 220 KB and a model of 230 KB failed to load.

A much larger model (still less than 1MB) could actually be loaded onto the board but that would involve converting the model into FlatBuffer using STM32Cube.AI and also recompiling and flashing the firmware to the board, leading to loss of the ease of access provided by OpenMV. The goal wasn't to run a large model anyway, hence this wasn't viable.

All the models were trained and tested on Kaggle kernels with TPU acceleration of 128 elements per core and 8 cores in parallel.

### B. Dataset Construction

Four datasets were used in this research work. The first dataset was the Sign MNIST dataset from Kaggle [20]. It was created from a set of 1704 uncropped color images taken in various backgrounds. The dataset creator used an ImageMagick pipeline to crop the images to hands-only, perform operations like gray-scaling and resizing to 28x28, and then creating 50+ variations to increase the quantity. The modification and expansion strategy involved filters Mitchell, Robidux, Catrom, Spline and Hermite, along with 5% random pixelation, +/- 15% brightness/contrast, and finally 3 degrees rotation. Due to the tiny size of the images, these modifications were sufficient to alter their properties like resolution and class separation in an interesting controlled manner. It had 24 classes for the 26 alphabets in English, barring J and Z, which require motion and cannot hence be detected from an image. The train set had 27,455 images and test set has 7172 images.

The second dataset was produced by modifying the Kaggle ASL dataset [21] having 87000 images of dimension 200x200 for 29 classes, the 26 English alphabets and 3 classes for Space, Nothing and Delete. Each class had 3000 images. This dataset has only 29 images for testing, the dataset creator mentioned it was to encourage the usage of real-world data. The classes J, Z, Space, Nothing and Delete were dropped. Thus, the resulting dataset had 72000 images for training and 24 for test.

The authors of this paper produced the third dataset using the target OpenMV H7 camera. The images taken were of 240x240 size and had approximately 400 images for each of the 24 classes, they were downscaled to 28x28 using the inter_area interpolation in OpenCV. The images were shuffled and split into 10% for training and 90% for testing. Some sample images are shown in Fig. 1 below.

The fourth dataset was the Kaggle ASL Alphabet Test dataset [22], created by Dan Rasband, in order to test the effectiveness of the models in real world scenarios with noisy backgrounds. It has 870 images, 30 images for each of the 29 classes, matching dataset [21]. Only 24 of those 29 classes are relevant for this research and hence 720 images were considered as the generalization dataset as the final test in this paper.

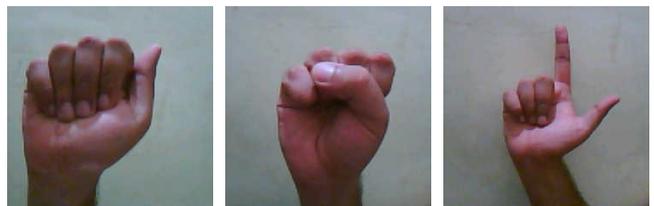

Fig. 1. Sample images from the third dataset for the letters A, S, L.

## IV. Experimental methodology

This section discusses the experimental steps data splitting, consequent hypothesis testing and comparison.

### A. Data Splitting

The training data from the first three datasets were all merged into one and had a little over $10^5$ images. The test set was constructed by merging the test sets from the three datasets and had about 23600 images, with the majority coming from the third dataset itself, that was created with the OpenMV camera.

### B. Proposed CNN Architecture

After a lot of experimentation with models of various architectures, keeping the size constraints of the deployment environment in mind, the architecture in Fig. 2 was found to be the most optimal, having 171,032 trainable parameters. Since each weight was quantized to int8 and was 1 byte, the complete size of the quantized model and weights was 185KB. Netron visualization of the architecture is shown alongside in Fig. 2a.

### C. Training Specifications

The above architecture was designed in TensorFlow 2.3. Categorical cross-entropy was chosen as the loss function and is given by the formula

$$Loss = -\sum_{c=1}^{M} y_{o,c} \ln(p_{o,c}) \qquad (1)$$

where M is the number of classes or 24, $y$ is the binary indicator if $c$ is the correct prediction for observation $o$, and $p$ is the predicted probability for observation $o$ being of class $c$.

The chosen optimizer was Adam with the default learning rate of 0.001, first moment exponential decay of 0.9, second moment exponential decay of 0.999, and an epsilon of 1e-07 for numerical stability. The callbacks used were ReduceLROnPlateau and ModelCheckPoint. The former was used to reduce the learning rate by a factor of 0.2 if the validation accuracy didn't improve after 5 epochs and the latter was used to save the best weights to a file on disk. These settings were kept constant for all the experiments to maintain uniformity.

### D. Post-training Procedure

Full Integer Quantization was carried out post-training using TensorFlow Lite, where all the weights and biases of the model were converted from float32 to int8 precision [128,127], using a representative dataset that measures the dynamic range of activations and weights. This representative dataset is a subset of the test set and is used by the converter to understand the range of inputs that shall be seen by the network in deployment. Quantization makes the model smaller and more efficient. The quantized model is shown in Fig. 2b.

### E. Evaluation

TensorFlow Lite Interpreter was used to evaluate the quantized model on the host computer. The quantized model was loaded to the TF Lite Interpreter and the test set was used to evaluate the metrics.

For evaluation at the edge with real world images, the quantized model was added to the OpenMV board using an SD card and loaded onto the RAM using MicroPython functions. A test script captured camera images at 240x240 resolution, rescaled to 28x28 and normalized them before feeding to the model. The images and predictions were saved on the SD card. This process was used for speed and performance testing.

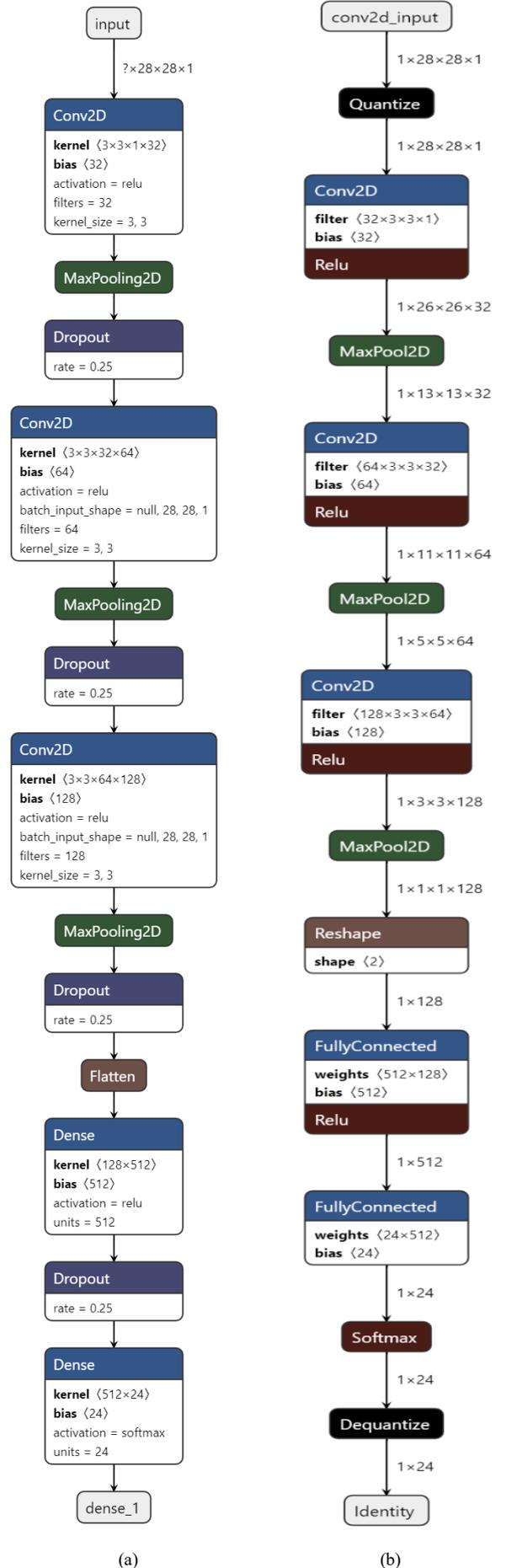

Fig. 2. Visualization of (a) float32 model and (b) quantized int8 model.

*F. Comparison between two augmentation methods*

The same architecture was trained using two augmentation techniques. For model 1, the train set was augmented using standard augmentation techniques like +/- 20° rotation, random cropping with 20x20 crops followed by resizing to 28x28, minor changes in contrast.

For model 2, the images were resized from the original 240x240 to 28x28 using the five interpolation strategies in OpenCV namely, inter_nearest, inter_linear, inter_area, inter_cubic, and inter_lanczos4 for augmentation.

The total size of the augmented train set and the test sets were kept constant for both the models. Then the effects of these two augmentation methods on the float32 model and the quantized model were carefully analyzed.

V. RESULTS AND DISCUSSION

This section discusses the findings and analysis with possible conclusions from the training and inference of the models with two different augmentation methods. It also discusses some strategies to help researchers take better decisions in these scenarios.

*A. Model 1*

Following the methodology described in Sec. IV A-C, the training accuracy rose almost uniformly and reached 92.96% within 40 epochs. The testing accuracy was 98.53%. Fig. 3 below shows the accuracy and loss curves.

Model 1 was then quantized to int8 and the accuracy plummeted to 95.283%, a drop of about 3.25%. Figs. 4 and 5 show the confusion matrices of the float32 and the int8 predictions respectively. Fig. 6 shows some sample predictions from Model 1 with both the precision settings.

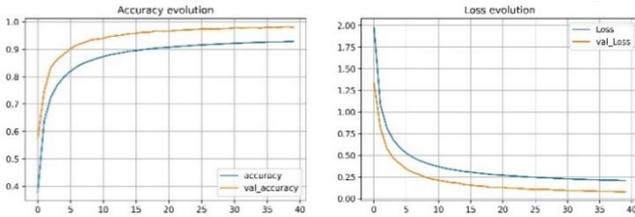
Fig. 3. Accuracy and Loss curves for Model 1

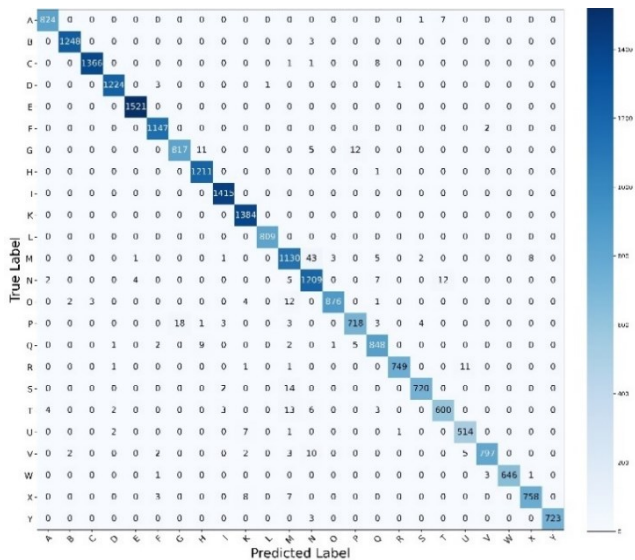
Fig. 4. Confusion matrix for Model 1 with float32 precision

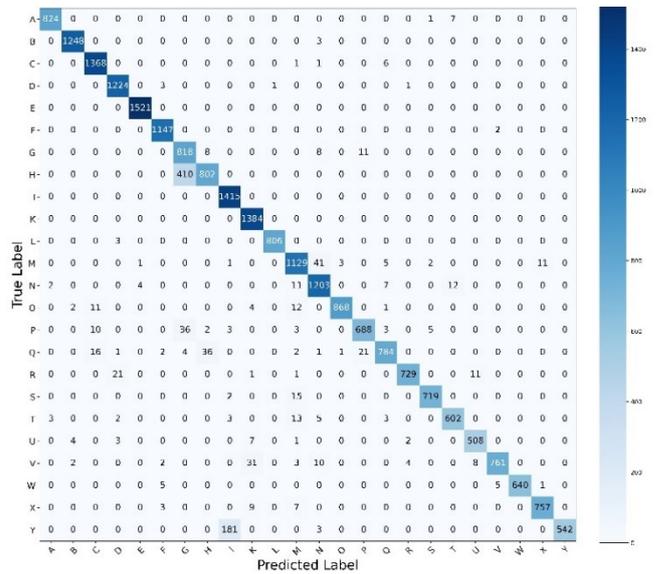
Fig. 5. Confusion matrix Model 1 with int8 precision

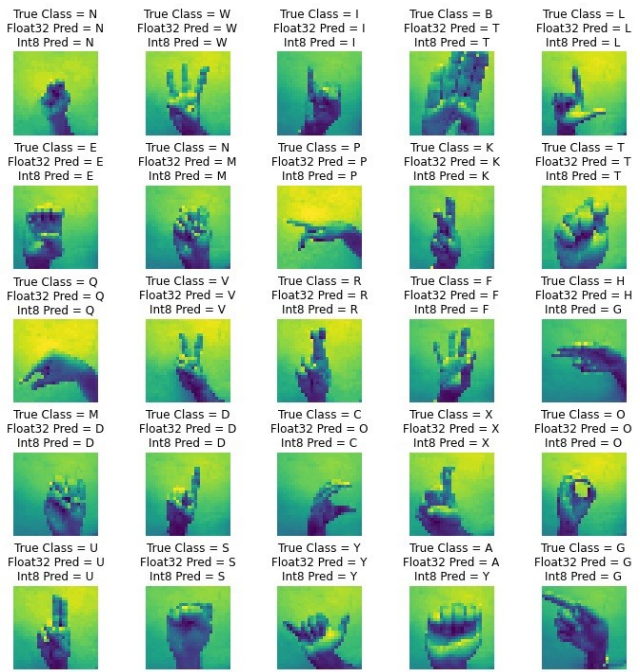
Fig. 6. Sample predictions of some images with Model 1

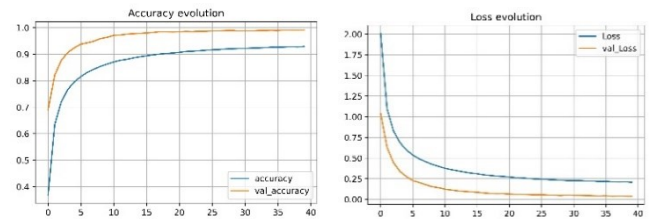
Fig. 7. Accuracy and Loss curves for Model 2.

*B. Model 2*

Similarly, for Model 2, the training accuracy rose almost uniformly and reached 92.63% within 40 epochs. The testing accuracy was 99.02%. Fig. 7 shows the accuracy and loss curves during training.

Figs. 8 and 9 show the confusion matrices of the float32 and the int8 predictions respectively. Fig. 10 shows some sample predictions from Model 2 with both the precision settings. When Model 2 was quantized to int8, the accuracy dropped to 98.84%, but the difference was a negligible 0.18%. Also, the quality of the predictions in int8 appear to be much better though the float32 predictions seem to be equivalent.

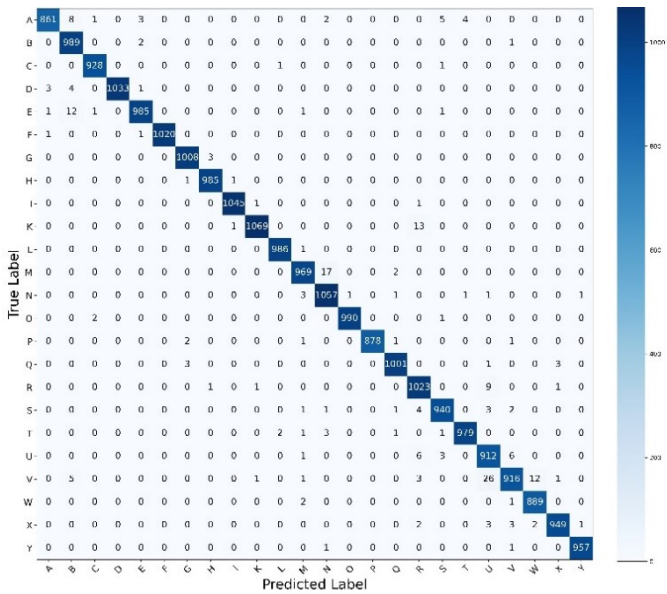

Fig. 8. Confusion matrix for Model 2 with float32 precision

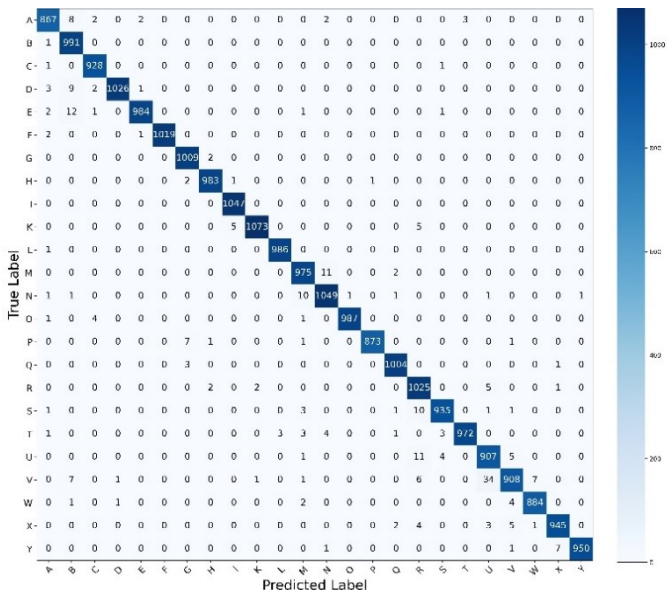

Fig. 9. Confusion matrix for Model 2 with int8 precision

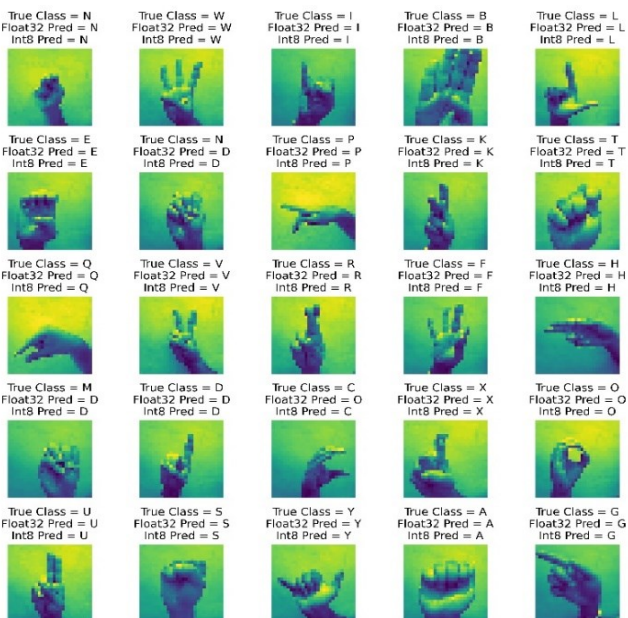

Fig. 10. Sample predictions of some images with Model 2.

*C. Discussion*

The images used in Fig10 are the same as those in Fig. 6 and may be used to draw some insights about both the models. However, caution must be used as 25 images out of 23600 barely give deep insights into model behavior and comparing the confusion matrices might thus be more worthwhile for inferential statistics. Nevertheless, it can be observed from the last row in Figs. 6 and 10, where the letter a is mispredicted in int8 in Model 1, but correctly predicted in int8 in Model 2.

Also, in Model 1, the int8 model mislabels 181 Y's as I's and 410 H's as G's (see Fig. 5), but the int8 Model 2 doesn't make any such error. In ASL, G, H and Y, I are very similar and model 2 performing better here is a clear sign that it is less affected by the quantization stage. Thus, the first conclusion is,

*"Interpolation augmentation appears to reduce the drop in accuracy during quantization in hand sign classification."*

During our experiments, the model inputs and outputs were kept first as int8 and then float32 and the second conclusion from our observations is,

*"Keeping inputs and outputs in float32 gives the best results."*

Another test that was done to find the ability of our models to generalize was to test it against the ASL Alphabet Test dataset [22]. [18] describes 'generalization error' as the difference between train and test sets, however, since usually training and test sets have elements coming from the same distribution (here it could be same person's hand or the same camera), and thus have similar underlying features the model might perform well on the train and test sets, but not on a generalization dataset from a completely different distribution. Bheda and Radpour [14] also noted this as an issue and this is a pretty common problem in TinyML, as it's not possible to learn too many features due to small model sizes. The results of our test of the int8 versions of Models 1 and 2 are in Table 1, along with the train and test accuracies after 40 epochs as well.

Table 1. Train, Test, and Generalization Accuracy after 40 epochs.

|  | Train acc. (%) | Test acc. (%) Float32 model | Test acc. (%) Int8 model | Generalization Acc. (%) |
|---|---|---|---|---|
| Model 1 | 92.96 | 98.53 | 95.28 | 70.97 |
| Model 2 | **92.63** | **99.02** | **98.84** | **74.58** |

Some images from the generalization dataset are shown in Fig. 11, exhibiting a variety of noisy backgrounds, but yet Model 2 turns out to be quite effective. This sort of testing on images from a different distribution is often not considered in research, but is of pivotal importance to understand generalization. From these observations, the third conclusion is,

*"Interpolation augmentation seems to improve classification generalization."*

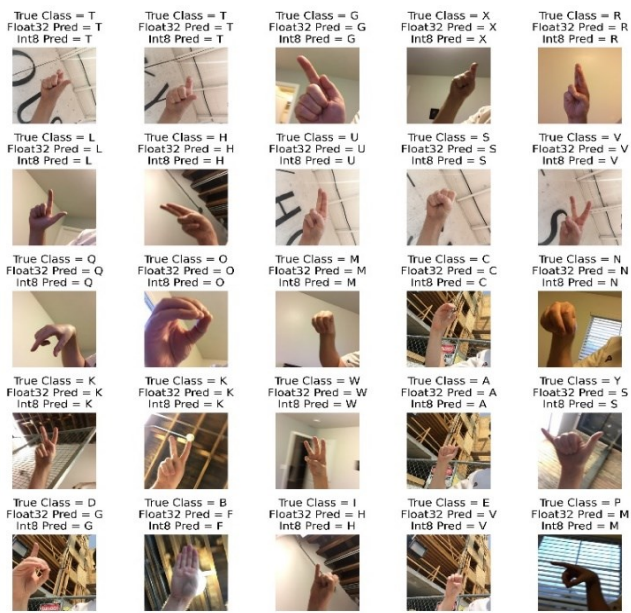

Fig. 11. Sample predictions of some images on the generalization dataset.

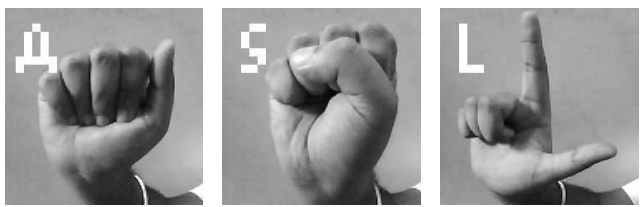

Fig. 12. Sample on-device predictions with Model 2.

Model 2 performed ASL classification on-device at a speed of 20 frames per second without being connected to a computer. Some experimental glimpses are in Fig. 12.

## VI. CONCLUSION

In this research, an extremely efficient, real-time, well-generalizable CNN based solution for ASL fingerspelling recognition has been implemented on an ARM Cortex-M7 powered OpenMV H7 board, which is just 185 KB in size and inference speed is 20fps. It has 98.84% test accuracy and 74.58% generalization accuracy. It has been shown that interpolation which is a normal step in image resizing, might be used to reduce accuracy drop during full integer quantization, simultaneously improving model generalizability, and that model performance is best if inputs and outputs are in float32.

The method proposed in this paper is highly generalizable and is applicable for any ARM microcontroller, and can also be easily ported to other architectures. Future avenues of research include improving the generalization accuracy by training it on data from more diverse sources and working on the hardware required to realize the vision of having a device, similar to a watch that is portable, easily affordable and can identify ASL fingerspelling. More work can also be done to identify the more complicated hand signs involving word level vocabulary and non-manual features which require identification of facial expressions, mouth tongue and body poses too.

This research thus makes ASL much more easily accessible, fostering equity and inclusivity of ASL users across the globe, besides making important observations about reducing accuracy drop during quantization, and model generalizability through use of interpolation augmentation.